\pdfoutput=1

\documentclass[11pt]{article}

\usepackage[dvipsnames, svgnames, x11names]{xcolor}
\usepackage{tikz}
\usepackage{pgfplots}
\usepackage{xcolor}
\usepackage{graphicx}
\usepackage{subfigure}
\usepackage{emnlp2021}
\usepackage{bookmark}
\usepackage{times}
\usepackage{latexsym}

\usepackage[T1]{fontenc}

\usepackage[utf8]{inputenc}
\usepackage{hyperref}
\usepackage{microtype}
\usepackage{amssymb}
\usepackage{pifont}
\newcommand{\xmark}{\ding{55}}%
\title{The NiuTrans Machine Translation Systems for WMT21}

\author{Shuhan Zhou\textsuperscript{1}, Tao Zhou\textsuperscript{1}, Binghao Wei\textsuperscript{1}, Yingfeng Luo\textsuperscript{1}, Yongyu Mu\textsuperscript{1},\\
        \bf{Zefan Zhou\textsuperscript{1}, Chenglong Wang\textsuperscript{1}, Xuanjun Zhou\textsuperscript{1}, Chuanhao Lv\textsuperscript{1}, Yi Jing\textsuperscript{1},}\\
        \bf{Laohu Wang\textsuperscript{1}, Jingnan Zhang\textsuperscript{1}, Canan Huang\textsuperscript{1}, Zhongxiang Yan\textsuperscript{1},}\\
        \bf{Chi Hu\textsuperscript{1}, Bei Li\textsuperscript{1}, Tong Xiao\textsuperscript{1,2} and Jingbo Zhu\textsuperscript{1,2}}\\
        \textsuperscript{1}NLP Lab, School of Computer Science and Engineering, Northeastern University\\
        \textsuperscript{2}NiuTrans Research, Shenyang, China\\
        \ttfamily{zhoushuhan199710@163.com,zhoutao\_neu@outlook.com}\\
        \ttfamily{\{xiaotong,zhujingbo\}@mail.neu.edu.cn}
        }

\begin{document}
\maketitle
\begin{abstract}

  This paper describes NiuTrans neural machine translation systems of the WMT 2021 news translation tasks. We made submissions to 9 language directions, including English$\leftrightarrow$$\{$Chinese, Japanese, Russian, Icelandic$\}$ and English$\rightarrow$Hausa tasks. Our primary systems are built on several effective variants of Transformer, e.g., Transformer-DLCL, ODE-Transformer. We also utilize back-translation, knowledge distillation, post-ensemble, and iterative fine-tuning techniques to enhance the model performance further.

\end{abstract}

\section{Introduction}

Our NiuTrans team participated in the WMT 2021 news translation shared tasks, including English$\leftrightarrow$Chinese (EN$\leftrightarrow$ZH), English$\leftrightarrow$Japanese (EN$\leftrightarrow$JA), English$\leftrightarrow$Russian (EN$\leftrightarrow$RU), English$\leftrightarrow$Icelandic (EN$\leftrightarrow$IS) and English$\rightarrow$Hausa (EN$\rightarrow$HA), nine submissions in total. All of our systems were built with constrained data sets. We adopt some effective models and useful methods, which have been witnessed the success in previous papers \cite{wang-etal-2018-niutrans,li-etal-2019-niutrans,zhang-etal-2020-niutrans,meng-etal-2020-wechat,wu-etal-2020-tencent,chen-etal-2020-facebook,yu-etal-2020-deepmind,wu-etal-2020-volctrans,wei-etal-2020-hw}. 

To enhance the performance of the single model, we choose pre-normalized Transformer-DLCL \cite{wang-etal-2019-learning-deep} and ODE-Transformer \cite{DBLP:journals/corr/abs-2104-02308} as the backbone. All systems are built upon the relative position representation \cite{shaw-etal-2018-self} due to its strong performance when models are deep \cite{li-etal-2020-shallow}. For the system combination, we adopt the post-ensemble \cite{kobayashi-2018-frustratingly} to find the most similar hypothesis among several ensemble outputs, which could be regarded as a reranking technique without pre-training. Previous works have emphasized the importance of diversity when building ensemble systems. Besides the architecture diversity, we also adopt iterative ensemble knowledge distillation leveraging the source-side monolingual data to enlarge the diversity. More details please refer to \cite{li-etal-2019-niutrans}. 

Our data preparation pipeline consists of three-fold: (\romannumeral1) For the data filtering. We use a stricter cleaning process than last year \cite{zhang-etal-2020-niutrans}. Details will be discussed in Section \ref{section2_2}. (\romannumeral2) For the data augmentation, both iterative back-translation \cite{sennrich-etal-2016-improving} method, and iterative knowledge distillation \cite{DBLP:journals/corr/FreitagAS17} method are employed to take the full advantage of monolingual data provided by the WMT organization. In the back-translation stage, we leverage target-side monolingual sentences to generate source-side pseudo sentences and use a nucleus sampling \cite{DBLP:journals/corr/abs-1904-09751} decoding strategy to improve the generalization ability. Furthermore, we leverage in-domain source-side monolingual data by applying iterative knowledge distillation. (\romannumeral3) For data selection, it’s hard to find massive in-domain data for low-resource languages to train a neural language model, so we use a statistical n-gram language model (XenC toolkit3\footnote{https://github.com/antho-rousseau/XenC}) instead.

\begin{table*}[htbp]
  \centering
  \begin{tabular}{lcccccc}
  \hline
  \textbf{Model} & \textbf{Depth} & \textbf{Hidden Size}& \textbf{Filter Size} & \textbf{RPR} & \textbf{Batch size} & \textbf{update freq}\\
  \hline
  Transformer & 6 & 512 & 2048 & \xmark & 4096 & 1 \\
  Transformer (Pre-Norm) & 24 & 512 & 4096 & \checkmark & 2048 & 4 \\
  Transformer-DLCL& 25 & 512 & 4096 & \checkmark & 2048 & 4 \\
  Transformer-DLCL& 30 & 512 & 2048 & \checkmark & 2048 & 4 \\
  Transformer-DLCL& 30 & 512 & 4096 & \checkmark & 2048 & 4 \\
  ODE Transformer& 6 & 1024 & 4096 & \checkmark & 2048 & 8 \\
  ODE Transformer& 12 & 1024 & 4096 & \checkmark & 2048 & 8 \\
  \hline
  \end{tabular}
  \caption{\label{table1}
  The details of several model architectures we used.
  }
  \end{table*}

  Domain finetuning is quite essential to improve the translation system given a certain target domain. We use domain adaptation to migrate the models from the general domain to the news domain by iterative finetuning. After in-domain finetuning, we use multiple ensemble combinations by the post-ensemble method.

This paper is structured as follows: In Section \ref{section2}, we introduce several effective techniques, including data preprocessing, deeper and wider Transformer models, iterative back-translation, iterative knowledge distillation, fine-tuning and post-ensemble. In Section \ref{section3}, we show the experiment settings and report the experimental results of the validation set (newstest2020). Finally, we draw the conclusion in Section \ref{section4}.

\section{System Overview}
\label{section2}

\subsection{Data Preprocessing and Filtering}
\label{section2_2}

For word segmentation, we use different tools in six languages. English, Russian, Hausa and Icelandic sentences were segmented by Moses \cite{koehn-etal-2007-moses}, while Chinese and Japanese used NiuTrans \cite{xiao-etal-2012-niutrans} and MeCab\footnote{https://github.com/taku910/mecab} separately. Then BPE \cite{sennrich-etal-2016-neural} with 32K operations is used for five languages sides independently, except for 36K operations in Russian.

The quality of the parallel training data is crucial to the performance of the models, so we use rigorous data filtering scheme as the suggestion in \citet{zhang-etal-2020-niutrans}'s work. For most language pairs, rules are as follows:

\begin{itemize}

  \item Filter out sentences that contain long words over 40 characters or over 150 words.
  
  \item The word ratio between the source word and the target word must not exceed 1:3 or 3:1.
  
  \item Use Unicode to filter sentences with more than 10 other characters.
  
  \item Filter out the sentences which contain HTML tags or duplicated translations.

  \item In monolingual data, some sentences contain two or more sentences. We write a script to cut them into several sentences.
\end{itemize}

We use these rules to filter bilingual and monolingual data, detecting low-quality sentences with misalignment, translation errors, illegal characters, and missing translation.

\subsection{Model Architectures}

As shown in previous work \cite{li-etal-2019-niutrans,zhang-etal-2020-niutrans,meng-etal-2020-wechat}, deep Transformers bring significant improvements than the baseline on various machine translation benchmarks. In their work, the performance of the model was significantly improved by increasing the encoder depth. We keep the decoder depth unchanged as the brought benefit is marginal when the encoder is strong enough \cite{DBLP:conf/aaai/LiWLDXZZ21}.

Hence, we train two deep models in our experiment: Transformer DLCL \cite{wang-etal-2019-learning-deep} and ODE Transformer \cite{DBLP:journals/corr/abs-2104-02308} with a larger filter size. ODE Transformer is designed from the ordinary differential equations (ODE) perspective. Higher-order ODE solutions can gain fewer truncation errors, thus reducing the global error and improving the model performance. The details of several models we mainly experimented with are summarized in Table \ref{table1}.

In addition, we incorporate relative position representation (RPR) into the self-attention mechanism on both the encoder and decoder sides. Preliminary experiments demonstrate that only relative key information is enough, and we set the relative window size to 8.

\subsection{Large-scale Back-Translation}

\definecolor{ublue}{rgb}{0.152,0.250,0.545}
\definecolor{ugreen}{rgb}{0,0.5,0}
\definecolor{lgreen}{rgb}{0.9,1,0.8}
\definecolor{amber}{rgb}{1.0, 0.75, 0.0}
\definecolor{xtgreen}{rgb}{0.914,0.945,0.902}
\definecolor{lightgray}{gray}{0.85}
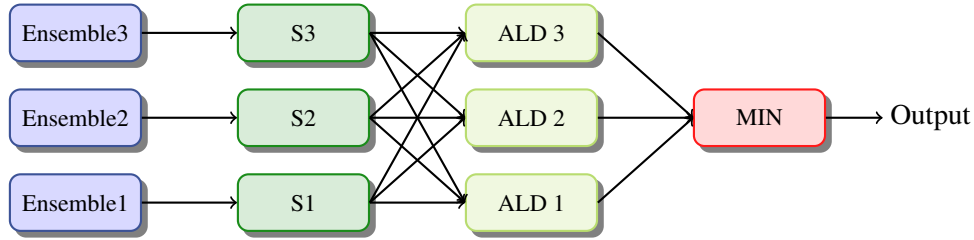
\begin{figure*}[htbp]
  \centering
  \tikzstyle{coder} = [rectangle,thick,rounded corners,minimum width=2.3cm,minimum height=1cm,text centered,draw=ublue!90,fill=blue!15]
\tikzstyle{t} = [rectangle,thick,rounded corners,minimum width=2.3cm,minimum height=1cm,text centered,draw=ugreen!90,fill=ugreen!15]
\tikzstyle{d} = [rectangle,thick,rounded corners,minimum 
width=2.3cm,minimum height=1cm,text centered,draw=YellowGreen!70,fill=YellowGreen!15]
\tikzstyle{shadow} = [rectangle,thick,rounded corners,minimum 
width=2.3cm,minimum height=1cm,text centered,draw=gray,fill=gray]

\begin{tikzpicture}[node distance = 0,scale = 0.75]
\tikzstyle{every node}=[scale=0.75]
\node(shadow1)[shadow]{};
\node(shadow2)[shadow,above of =shadow1,xshift=0cm,yshift=1.5cm]{};
\node(shadow3)[shadow,above of =shadow2, xshift=0cm,yshift=1.5cm]{};
\node(shadow4)[shadow,right of =shadow1,xshift=4cm,yshift=0cm]{};
\node(shadow5)[shadow,right of =shadow2,xshift=4cm,yshift=0cm]{};
\node(shadow6)[shadow,right of =shadow3,xshift=4cm,yshift=0cm]{};
\node(shadow7)[shadow,right of =shadow4,xshift=4cm,yshift=0cm]{};
\node(shadow8)[shadow,right of =shadow5,xshift=4cm,yshift=0cm]{};
\node(shadow9)[shadow,right of =shadow6,xshift=4cm,yshift=0cm]{};
\node(shadow10)[shadow,right of =shadow8,xshift=4cm,yshift=0cm]{};

\node(ensemble1)[coder,above of =shadow1,xshift=-0.1cm,yshift=0.1cm]{\large{Ensemble1}};
\node(ensemble2)[coder,above of =ensemble1,xshift=0cm,yshift=1.5cm]{\large{Ensemble2}};
\node(ensemble3)[coder,above of =ensemble2, xshift=0cm,yshift=1.5cm]{\large{Ensemble3}};

\node(text1)[t,right of =ensemble1, xshift=4cm,yshift=0cm]{\large{S1}};
\node(text2)[t,right of =ensemble2, xshift=4cm,yshift=0cm]{\large{S2}};
\node(text3)[t,right of =ensemble3, xshift=4cm,yshift=0cm]{\large{S3}};

\node(dis1)[d,right of =text1, xshift=4cm,yshift=0cm]{\large{ALD 1}};
\node(dis2)[d,right of =text2, xshift=4cm,yshift=0cm]{\large{ALD 2}};
\node(dis3)[d,right of =text3, xshift=4cm,yshift=0cm]{\large{ALD 3}};

\node(min)[d,right of =dis2, xshift=4cm,yshift=0cm,draw=red!90,fill=red!15]{\large{MIN}};
\node(output)[right of =min, xshift=3cm,yshift=0cm]{\Large{Output}};

\draw[->,thick](ensemble1.east)to(text1.west);
\draw[->,thick](ensemble2.east)to(text2.west);
\draw[->,thick](ensemble3.east)to(text3.west);

\draw[->,thick,fill=](text1.east)to(dis1.west);
\draw[->,thick,fill=](text1.east)to(dis2.west);
\draw[->,thick,fill=](text1.east)to(dis3.west);
\draw[->,thick](text2.east)to(dis1.west);
\draw[->,thick](text2.east)to(dis2.west);
\draw[->,thick](text2.east)to(dis3.west);
\draw[->,thick](text3.east)to(dis1.west);
\draw[->,thick](text3.east)to(dis2.west);
\draw[->,thick](text3.east)to(dis3.west);

\draw[->,thick,fill=](dis1.east)to(min.west);
\draw[->,thick](dis2.east)to(min.west);
\draw[->,thick](dis3.east)to(min.west);

\draw[->,thick](min.east)to(output.west);
\end{tikzpicture}
  \caption{Process of the post-ensemble method. S$n$ denotes the sentence generated by the $n$-th ensemble of models. ALD denotes the average Levinstein distance of a sentence with other sentences. For example, $ALD\, 1=\frac{1}{2}*(Levinstein\_distance(S1,S2)+Levinstein\_distance(S1,S3))$. Finally, we select the sentence of the smallest ALD.}
  \label{fig:1}
  \end{figure*}

Back-translation (BT) is an effective data augmentation technique to boost the performance of NMT models, which use monolingual data to generate pseudo-training parallel data. Back-translation is divided into three stages: 
\begin{itemize}
    \item Using bilingual parallel data to train a target-to-source intermediate ensemble of models.
    \item Utilizing the ensemble of reverse direction models to translate the target monolingual corpus into the source corpus.
    \item Training models with the bilingual parallel corpus and the synthetic parallel corpus together.
\end{itemize}

Select in-domain monolingual data during back-translation can significantly alleviate domain adaptation problems \cite{zhang-etal-2020-niutrans}. Our in-domain data consist of the test sets released in recent years and the News Commentary high-quality monolingual data. Due to insufficient data in the domain, we used a statistical method to select in-domain data, the XenC toolkit. Furthermore, to avoid the high ranking of short sentences, we choose the in-domain source side sentences according to the distribution of sentence tokens number in the previous years' test set.

For all tasks, we employ the beam search and Nucleus Sampling approaches to generate pseudo corpus and the scale of the pseudo corpus was about 1:1 to the real corpus.

\subsection{Iterative Knowledge Distillation}

Knowledge distillation (KD) has been proven to be a powerful technique to improve the performance of the student model by transferring knowledge from the teacher model \cite{li-etal-2019-niutrans,zhang-etal-2020-niutrans}. Here, we regard the ensemble models as a teacher model and single models as student models. Specifically, we first use the ensemble model to generate synthetic corpus in the forward direction. Then, we merge the synthetic parallel corpus with the bilingual parallel corpus to teach student models. And by searching for better model ensemble combinations, we can provide stronger teacher models for the next round of knowledge distillation. Our experiment found that the gap between the single model and the integrated model gradually narrowed as the iteration progressed. So for the nine tasks we participated in, two iterations of knowledge distillation deliver the best performance.

\subsection{Finetuning}

Domain adaptation plays an important role in improving the performance of the models. A practical method of domain adaptation is to train models on large-scale out-domain corpus and then fine-tune the models with in-domain corpus \cite{Luong2015StanfordNM}. For all tasks, we mainly reuse an iterative fine-tuning process \cite{zhang-etal-2020-niutrans} and use the development sets and the test sets of previous years as in-domain corpus.

It is worth noting that, in order to be consistent with the composition of the test set, we select parallel sentences pair from the previous development sets and test sets in which the source side is real and the target side is manually translated.  Moreover, we found that iterative fine-tuning can better improve the translation quality of the names of news organizations in the news field.

\subsection{Post-ensemble}

Ensemble learning is a technique widely used in several WMT shared tasks, which improves performance by using multiple single models. In neural machine translation, a practical method of the model ensemble is to combine the probability distribution on the target vocabulary of different models in each step of sequence prediction. Here, we adopted their method, which uses a greedy-based strategy to find a better combination of models on the development set. However, enumerating all combinations of candidate models is an inefficient and cumbersome way. 

In our ensemble experiments, we set the number of the ensemble to four and six. We observed that simply expanding the scale of the ensemble does not necessarily improve translation performance. Besides, brute force search for all models is costly and unrealistic. As the number of models increases, the ensemble easily exceeds the computer capacity limit. Therefore, for all tasks, we finally search for four single models as an ensemble.

In addition, we use a simple but effective unsupervised ensemble method, post-ensemble, which uses a clustering method to select a majority-like output from multiple ensembles. As shown in the figure \ref{fig:1}, we first choose several ensemble combinations composed of different models to obtain more diversity. Then we use these ensembles to generate multiple sentences, respectively. Next, we calculate the Levinstein distance between each sentence, and finally, we select the sentence of the smallest average Levinstein distance with other sentences.

For more detailed content, please refer to the original paper \cite{kobayashi-2018-frustratingly}. This technology can further improve the performance of the system based on ensemble learning.

\section{Experiment}
\label{section3}

\subsection{Experiment Settings}

The implementation of our models is based on Fairseq \cite{ott-etal-2019-fairseq}. All models were trained on 8 RTX 2080Ti GPUs. We selected the pre-norm Transformer-base as the baseline for all tasks and enhanced our deep or wide models by enlarging the model depth and the hidden size, respectively. We used Adam optimizer \cite{kingma2014adam} with $\beta_1 = 0.9$, $\beta_2 = 0.997$ during training. As suggested in \citet{ott-etal-2018-scaling} and \citet{wang-etal-2019-learning-deep}'s work, models with larger capacities tend to perform much better within large batch size and learning rate. Due to the high GPU memory consumption, accumulated gradients every two steps where each batch contains 2048 tokens. Training for 15 epochs is sufficient for most tasks, and models have shown convergence in validation perplexity. The max learning rate and warmup step were set to 0.002 and 8000 for deep models, and 0.0016 and 16000 for deep and wide models, e.g., Transformer-DLCL, whose hidden dimension is 768. All the dropout probabilities were set to 0.1, including the residual dropout, attention dropout, and the ReLu dropout. We also used FP16 mix-precision training to accelerate further the training process with almost no loss in BLEU.

\setlength{\tabcolsep}{3.6mm}{
\begin{table*}[htbp]
  \centering
  \begin{tabular}{lccccc}
  \hline
  \textbf{System} & \textbf{EN$\rightarrow$ZH} & \textbf{ZH$\rightarrow$EN} & \textbf{EN$\rightarrow$JA} & \textbf{JA$\rightarrow$EN} & \textbf{EN$\rightarrow$HA} \\
  \hline
  Baseline & 41.9 & 30.1 & 34.5 & 21.4 & 10.9\\
  DLCL30-RPR & 42.8 & 31.0 & 35.6 & 21.6 & 11.9\\
  +Iteratively BT & 46.5 & 33.3 & 38.4 & 21.7 & 16.5\\
  +Iteratively KD& 47.4 & 35.0  & 41.8 & 25.9 & 18.2\\
  +Fine-tune& 47.8 & 37.0 & 42.0 & 26.4 & - \\
  \hline
  +Ensemble& 48.8 & 37.2 & 42.7 & 27.4 & 18.5 \\
  +Post-ensemble \& Post edit& 49.0 & 37.5 & 43.6 & 27.4 & - \\
  \hline
  \end{tabular}
  \caption{\label{table2}
  BLEU evaluation results on the WMT 2020 EN$\leftrightarrow$ZH, EN$\leftrightarrow$JA test sets and WMT 2021 EN$\rightarrow$HA development sets.
  }
\end{table*}}

\subsection{EN$\leftrightarrow$ZH}

For EN$\leftrightarrow$ZH tasks, the training data consists of ParaCrawl, News Commentary v16, WikiMatrix, UN Parallel Corpus V1.0, and the CCMT Corpus. We regarded the newstest2019 as the valid set and the newstest2020 as the test set to tune the hyper-parameters. After filtering the data, we sampled the top 12 and 20 million data according to the XenC score as the bilingual dataset. For the ZH$\rightarrow$EN task, we used 12 and 20 million data to train the baseline model, respectively, and found that the model trained by 12 million data is 1 and 1.2 BLEU point higher than the model trained by 20 million data in the valid and test set. We found that the data quality of the bottom 8 million is lower and also selected the 12 million data as our training data.

During the first-step back-translation, we sampled 8 million monolingual data from the combination of News crawl, News Commentary, News discussions, and News crawl. Then we used the baseline model to generate the hypotheses via the beam search strategy as the pseudo dataset. In the second-step back-translation, we utilized the same amount of pseudo data while using nucleus sampling, whose p is 0.9. For ZH$\rightarrow$EN and EN$\rightarrow$ZH, we got BLEU improvements of 1.8 and 2.9 in the first back-translation and further BLEU improvements of 0.5 and 0.8 in the second back-translation, respectively.

In addition, we implemented knowledge distillation twice to iteratively enhance the single model with the ensemble outputs. The main goal is to make the single student mimic the behavior of the ensemble models, thus obtaining stronger ensemble teachers in the next step. We used the test sets in previous years as in-domain data in EN$\rightarrow$ZH and EN$\rightarrow$ZH directions respectively, and we used the XenC tool to sample 3 million from the large scale monolingual data based on in-domain data. Then we used the best ensemble of models to construct pseudo data by decoding them and merge them to the original training data to continue training for each model. We got BLEU improvements of 1.1 and 0.6 in the first knowledge distillation and further BLEU improvements of 0.6 and 0.3 in the second knowledge distillation in ZH$\rightarrow$EN and EN$\rightarrow$ZH. 

After knowledge distillations, we used the newstest2017-2019 to fine-tune our models for five epochs with the 0.0001 learning rate and got 2 and 0.4 BLEU improvements in EN$\rightarrow$ZH and EN$\rightarrow$ZH directions, respectively. In the final stage, we add newstest2020 to the fine-tuning data. Finally, we searched for the best five combinations of 4 out of 12 models for post-ensemble to ensure the diversity of the models. Based on the ensemble method, post-ensemble further brought us +0.2 and +0.3 BLEU in ZH$\rightarrow$EN and EN$\rightarrow$ZH directions. Our main results showed in table \ref{table2}, we find that iterative back-translation, iterative knowledge distillation, and iterative fine-tune are effective methods to get significant improvements.

\subsection{EN$\leftrightarrow$JA}

\setlength{\tabcolsep}{5mm}{
\begin{table*}[htbp]
  \centering
  \begin{tabular}{lcccc}
  \hline
  \textbf{System} & \textbf{EN$\rightarrow$RU} & \textbf{RU$\rightarrow$EN} & \textbf{EH$\rightarrow$IS} & \textbf{IS$\rightarrow$EN} \\
  \hline
  Baseline & 22.0 & 35.6 & 20.9 & 28.4 \\
  ODE big6-RPR & 22.7 & 36.8 & 22.4 & 30.5 \\
  +Iteratively BT & 23.0 & 38.2 & 28.5 & 34.9 \\
  +Iteratively KD & 23.3 & 38.9 & 30.7 & 36.0 \\
  +Fine-tune& 24.4 & 39.4 & - & - \\
  \hline
  +Ensemble& 24.8 & 39.9 & 31.2 & 36.4 \\
  \hline
  \end{tabular}
  \caption{\label{table3}
  BLEU evaluation results on the WMT 2020 EN$\leftrightarrow$RU test sets and WMT 2021 EN$\leftrightarrow$IS development sets.
  }
\end{table*}}

For EN$\leftrightarrow$JA tasks, we chose ParaCrawl v7.1, News Commentary v16, WikiMatrix, Japanese-English Subtitle Corpus, The Kyoto Free Translation Task Corpus, TED Talks total of six parallel data corpora about 17.5 million. For the ParaCrawl v7.1, we only selected 8.5 million data according to the score of sentences provided by the dataset. We chose all of News Crawl and News Commentary and 12 million data sampled from Common Crawl for the Japanese monolingual data. After merging corpora into training data, we found that there were many-to-one situations in both the target side and the source side. Therefore, we sorted sentences and calculated the Levenshtein ratio of two adjacent sentences to remove duplication sentences. We applied this method to all version data before training models and removed 10 percent of the total data. We randomly selected one out of many sentences in which Levenshtein ratios are greater than or equal to 0.9.

We also implemented tagged back-translation, which brought us +2.8 BLEU in the EN$\rightarrow$JA task. In addition, beam search and nucleus sampling were used to generate two parts of translations to increase data diversity, and each part contains 12 million data. An interesting phenomenon is that back-translation is useful for EN$\rightarrow$JA task while knowledge distillation is helpful for JA$\rightarrow$EN task. We suspect this is because the domain of Japanese monolingual more fits the field of the test set.

We also implemented knowledge distillation and fine-tuned iteratively. During the knowledge distillation phase, we used FDA\footnote{https://github.com/bicici/FDA} and XenC to select monolingual data more like newstest2020 and generated pseudo data by using both post-ensemble and ensemble methods. During the fine-tuning phase, we used the WMT 2020 valid set and opposite direction test set. After performance stopped increasing at the second fine-tune, we utilized the best ensemble models to regenerate pseudo data by back-translation and knowledge distillation. Then, we retrained multiple deep models. Finally, we put all models together to greedy search for the best combination of 13 models. And this method brought us +0.7 BLEU in JA$\rightarrow$EN task. Our main results are shown in table \ref{table2}.

\subsection{EN$\leftrightarrow$RU}

For EN$\leftrightarrow$RU tasks, we used only two parallel datasets, including ParaCrawl v8 and News Commentary. After the data filter, about 12M sentence pairs were left to build our system. Additionally, we set the merge operations of BPE to 36K. 

We also used iterative back-translation, iterative knowledge distillation, and fine-tuned to enhance the model. During the back­-translation, English monolingual data is the same as the EN$\leftrightarrow$JA part, and Russian monolingual data sources consist of News Crawl and News Commentary. During the knowledge distillation, we used FDA to select 4 million sentence pairs from the monolingual dataset according to the newstest2020 and newstest2019. Then we merged them with the official development set to continue training our models for five epochs. After KD, we used the newstest2019 and newstest2018 to fine-tune our models for five epochs with the 0.0001 learning rate and got 1.1 and 0.5 BLEU improvements in EN$\rightarrow$RU and RU$\rightarrow$EN.

The detailed and full results can be described in Table \ref{table3}. Iterative BT, KD, and fine-tune are still very effective and improved 2.8 and 4.3 compared with the base model in EN$\rightarrow$RU and RU$\rightarrow$EN tasks, respectively.

\subsection{EN$\leftrightarrow$IS}

The process of EN$\leftrightarrow$IS tasks is similar to EN$\rightarrow$HA task but more complicated. Concretely, we used four parallel datasets, including ParaCrawl v7.1, Wiki Titles v3, WikiMatrix, and ParIce. After the data filtering, about 5.5 million sentence pairs were left to build the baseline system. The experimental results are listed in Table \ref{table3}. We obtained significant improvements of 6.1 and 4.4 BLEU in EN$\rightarrow$IS and IS$\rightarrow$EN directions, respectively.


Then we implemented iterative KD two times and sampled 3 million in-domain source data according to WMT2021 development sets. Table \ref{table3} shows that it’s a very effective method to get 2.2 and 1.1 improvements. Furthermore, we fine-tuned models iteratively twice to transfer the knowledge into the target domain. Due to implementing two ensemble combinations to decode sentences, the model ensemble still gained 0.7 and 0.8 improvements.

\setlength{\tabcolsep}{1mm}{
\begin{table}
  \centering
  \begin{tabular}{lclc}
  \hline
  \textbf{Task} & \textbf{Submission} & \textbf{Task} & \textbf{Submission}\\
  \hline
  EN$\rightarrow$ZH & 35.8 & EN$\rightarrow$RU & 28.4  \\
  ZH$\rightarrow$EN & 31.9 & RU$\rightarrow$EN & 41.8 \\
  EH$\rightarrow$JA & 46.2 & EH$\rightarrow$IS & 30.6 \\
  JA$\rightarrow$EN & 27.2 & IS$\rightarrow$EN & 39.2\\
  EH$\rightarrow$HA & 19.7 & - & - \\
  \hline
  \end{tabular}
  \caption{\label{table4}
  Our final submission results in nine tasks.
  }
\end{table}}

\subsection{EN$\rightarrow$HA}

In the EN$\rightarrow$HA direction, we used ParaCrawl v8, Khamenei corpus, and English-Hausa Opus corpus three data sets, obtaining 1.43M parallel data after cleaning. We collected News crawl, Extended Common Crawl, and Common Crawl for the monolingual data, resulting in 5.7M Hausa monolingual data. Considering the insufficient scale of Hausa, we used all monolingual data in each round of back-translation. The implementation details of iterative knowledge distillation and back-translation are almost the same as the EN$\leftrightarrow$ZH tasks.

Table \ref{table2} summarized the results. We can observe that the wide and deep models were still effective in low-resource language pairs. Through the back-translation and knowledge distillation techniques, we gain 4.6 and 1.7 BLEU improvements, respectively.
\subsection{Submission Results}

The results we finally submitted are shown in table \ref{table4}. We participated in nine tasks this year. On the whole, all of our systems performed competitively, especially in EH$\rightarrow$IS and RU$\rightarrow$EN directions. 

Through all the experimental results, we found that different methods perform differently on nine tasks. Among them, iterative BT is effective for almost all tasks, except for the JA$\rightarrow$EN task. Iterative KD performs better for EN$\leftrightarrow$ZH, EN$\leftrightarrow$JA and EH$\leftrightarrow$IS tasks, while fine-tune is more suitable for ZN$\rightarrow$EN and EN$\rightarrow$RU tasks.

\section{Conclusion}
\label{section4}

This paper introduced our submissions on WMT21 nine tasks. Our main exploration is using a new effective architectures ODE Transformer and utilizing post-ensemble technology to enhance the system. And we experimented with iterative back-translation by different decoding strategies, iterative knowledge distillation, iterative fine-tuning, model ensembling and post-ensemble.

\bibliography{anthology,custom}
\bibliographystyle{acl_natbib}

\end{document}